\documentclass[10pt,conference]{IEEEtran}

\usepackage{cite}
\usepackage{amsmath,amssymb,amsfonts}
\usepackage{graphicx}
\usepackage[table]{xcolor}
\usepackage{booktabs}
\usepackage{url}
\usepackage[hidelinks]{hyperref}

\title{SCoRe: Clean Image Generation from Diffusion Models Trained on Noisy Images}

\author{
\IEEEauthorblockN{Yuta Matsuzaki}
\IEEEauthorblockA{
Kyushu University\\
Fukuoka, Japan\\
yuta.matsuzaki@human.ait.kyushu-u.ac.jp
}
\and
\IEEEauthorblockN{Seiichi Uchida}
\IEEEauthorblockA{
Kyushu University\\
Fukuoka, Japan\\
uchida@ait.kyushu-u.ac.jp
}
\and
\IEEEauthorblockN{Shumpei Takezaki}
\IEEEauthorblockA{
Kyushu University\\
Fukuoka, Japan\\
shumpei.takezaki@human.ait.kyushu-u.ac.jp
}
}
\begin{document}
\maketitle 
\begin{abstract}
Diffusion models trained on noisy datasets often reproduce high-frequency training artifacts, significantly degrading generation quality. To address this, we propose SCoRe (Spectral Cutoff Regeneration), a training-free, generation-time spectral regeneration method for clean image generation from diffusion models trained on noisy images. Leveraging the spectral bias of diffusion models, which infer high-frequency details from low-frequency cues, SCoRe suppresses corrupted high-frequency components of a generated image via a frequency cutoff and regenerates them via SDEdit. Crucially, we derive a theoretical mapping between the cutoff frequency and the SDEdit initialization timestep based on Radially Averaged Power Spectral Density (RAPSD), which prevents excessive noise injection during regeneration. Experiments on synthetic (CIFAR-10) and real-world (SIDD) noisy datasets demonstrate that SCoRe substantially outperforms post-processing and noise-robust baselines, restoring samples closer to clean image distributions without any retraining or fine-tuning.

\end{abstract}

\begin{IEEEkeywords}
Generative model, Diffusion model, Image generation
\end{IEEEkeywords}

\section{Introduction}
Diffusion models~\cite{diffusionmodel,DDPM} are widely used as generative models capable of producing high-quality images. In these models, a diffusion process gradually transforms real images into random noise, while image generation is achieved by learning the reverse (denoising) process. Owing to their strong generative performance, diffusion models have been successfully applied to a wide range of image generation tasks and have become a cornerstone of recent image generation research~\cite{10081412,10.1145/3626235,10419041}.

\begin{figure}[t]
    \centering
    \includegraphics[keepaspectratio, width=\linewidth]{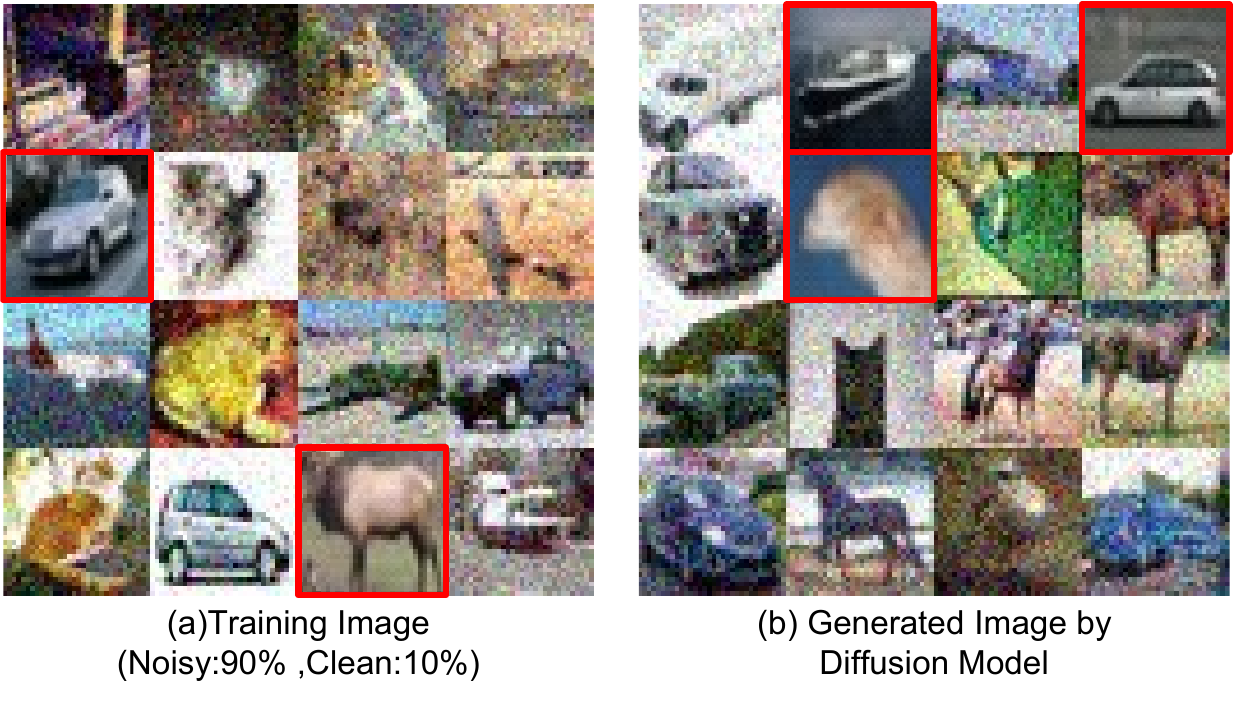}\\[-4mm]
    \caption{Impact of a dataset containing noisy training images on the generated results. Images in red boxes are clean examples; all others are noisy.}
    \label{noise_and_clean}
    \vspace{-5pt}
\end{figure}

However, when the training data contains noisy images, the quality of images generated by diffusion models often deteriorates. Fig.~\ref{noise_and_clean}~(a) shows an example dataset in which only 10\% of the training images are clean and the remaining 90\% are noisy, and Fig.~\ref{noise_and_clean}~(b) presents representative generation results obtained from such data. As illustrated, a large portion of the generated images is degraded. This phenomenon can be interpreted as the coexistence of clean and noisy image distributions in the training dataset, where the noisy distribution is dominant. As a result, the diffusion model primarily learns this dominant noisy distribution, leading to degraded generation quality.

Importantly, the mixed clean/noisy setting in Fig.~\ref{noise_and_clean} is not an artificial corner case. In practical data collection pipelines, large-scale image corpora are often aggregated from heterogeneous sources and acquisition conditions, where a substantial portion of samples inevitably contain noise or artifacts (e.g., sensor noise, low-light captures, compression, or resizing). At the same time, it is common to have only a limited subset of clean images obtained through manual curation, controlled acquisition, or costly verification procedures. Consequently, diffusion models are frequently trained on datasets in which the noisy mode dominates, even though some clean examples are available. This motivates generation-time approaches that can leverage pre-trained diffusion models to recover cleaner samples without requiring explicit noise modeling or retraining.

A straightforward approach to address this issue is to apply denoising as a post-processing step to the generated images. However, such post-processing often results in unnatural images. Classical denoising filters~\cite{Bilateral}, for example, tend to remove not only noise but also important image components such as textures. Similarly, neural-network-based denoising methods~\cite{Noise2Void} do not guarantee that the restored images faithfully follow the original data distribution, as they operate independently of the generative model.

\begin{figure*}[t]
    \centering
    \includegraphics[width=\linewidth]{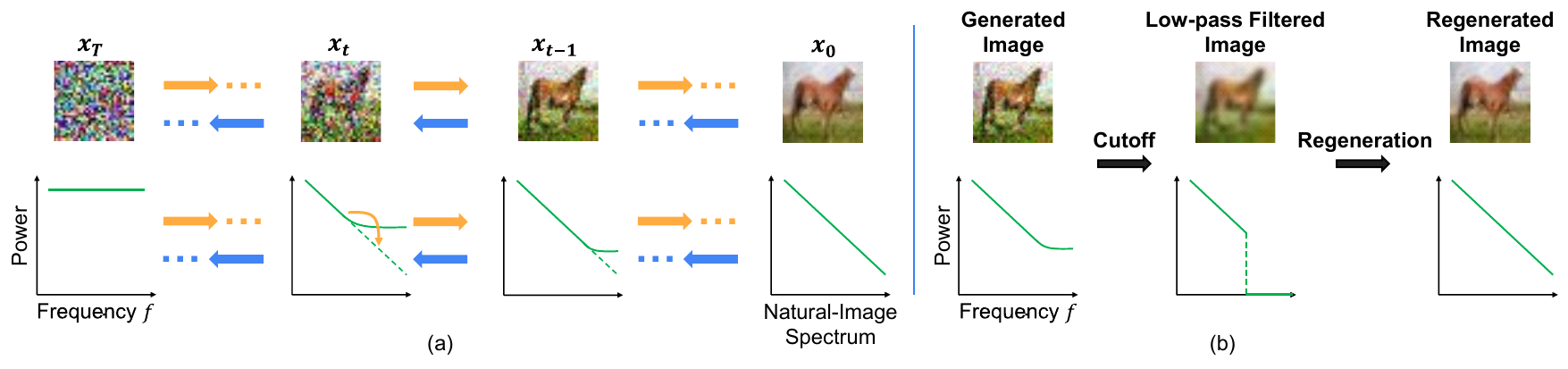}\\[-3mm]
    \caption{(a) Overview of the diffusion and reverse processes in terms of frequency components. (b) \textbf{SCoRe}: training-free, generation-time spectral regeneration that suppresses high-frequency components via a frequency cutoff and regenerates high-frequency details via SDEdit.}
    \label{diffusion_process_RAPSD}
    \vspace{-5pt}
    
\end{figure*}

In this work, we propose \textit{SCoRe (Spectral Cutoff Regeneration)}, a training-free, generation-time spectral regeneration method for clean image generation from diffusion models trained on datasets containing both clean and noisy images. Rather than modifying the training procedure or explicitly modeling the noise process, SCoRe intervenes only during sampling by suppressing corrupted high-frequency components and regenerating them under the diffusion-model prior. Since it requires no additional training or fine-tuning, SCoRe can be applied directly to pretrained diffusion models.

Our approach is motivated by a frequency-domain perspective of diffusion models. In many real-world images, semantic structures are largely captured by low-frequency components, whereas undesired noise predominantly appears in the high-frequency band. Moreover, diffusion models exhibit a characteristic tendency to infer higher-frequency components based on lower-frequency components during the generation process~\cite{spectral} as shown in Fig.~\ref{diffusion_process_RAPSD}~(a). Leveraging this property, we guide the generation results toward cleaner images by manipulating frequency components during sampling.

Specifically, as shown by Fig.~\ref{diffusion_process_RAPSD}~(b), we suppress the high-frequency components of a generated image using a cutoff operation and subsequently perform regeneration via SDEdit~\cite{SDEdit}. This procedure preserves low-frequency components while allowing only the high-frequency components to be re-estimated according to the diffusion model prior. Since the cutoff operation reduces the power of high-frequency components to near zero, the regeneration process starts from an initial state that is closer to clean high-frequency statistics than to noisy ones. Consequently, the regenerated high-frequency components are more likely to follow a cleaner distribution.

To evaluate the effectiveness of the proposed method, we conduct image generation experiments on multiple datasets. Specifically, we consider synthetic noise applied to CIFAR-10 as well as real-world noise in the SIDD dataset. Experimental results demonstrate that our proposed generation-time spectral regeneration method enables the generation of clean images, even when the diffusion model is trained on datasets containing a large proportion of noisy images.

\section{Related Work}
\noindent\textbf{Post-processing Denoising:}\ 
A common approach involves applying denoising filters to generated images. Classical methods, such as bilateral filters~\cite{Bilateral} and BM3D~\cite{BM3D}, smooth images locally, while learning-based methods like Noise2Void~\cite{Noise2Void} and Noise2Noise~\cite{Noise2Noise} remove noise statistically without clean supervision. However, since these post-processing methods operate independently of the generative model, they often inadvertently remove high-frequency structural details along with the noise, degrading the fidelity and naturalness of the results.

\noindent\textbf{Noise-Robust Generative Models:}\ To handle noisy training data, robust training frameworks have been proposed. For GANs, AmbientGAN~\cite{AmbientGAN} addresses known noise distributions, while NR-GAN~\cite{NR-GAN} and BNCR-GAN~\cite{BNCR-GAN} extend this to unknown or diverse noise types. Similarly, for diffusion models, Ambient Diffusion~\cite{AmbientDiffusion} and its variants~\cite{AmbientDiffusion-2,AmbientDiffusion-3} incorporate corruption operators into the training process. 

While effective, these learning-based approaches typically require embedding specific noise constraints into the training objective, necessitating noise-specific retraining and incurring high computational costs. In contrast, our proposed method focuses solely on the sampling process of a pre-trained diffusion model. It generates clean images without any additional training or prior knowledge of the noise distribution, making it a highly practical, training-free solution.

\begin{figure*}[t]
    \centering
    \includegraphics[width=\linewidth]{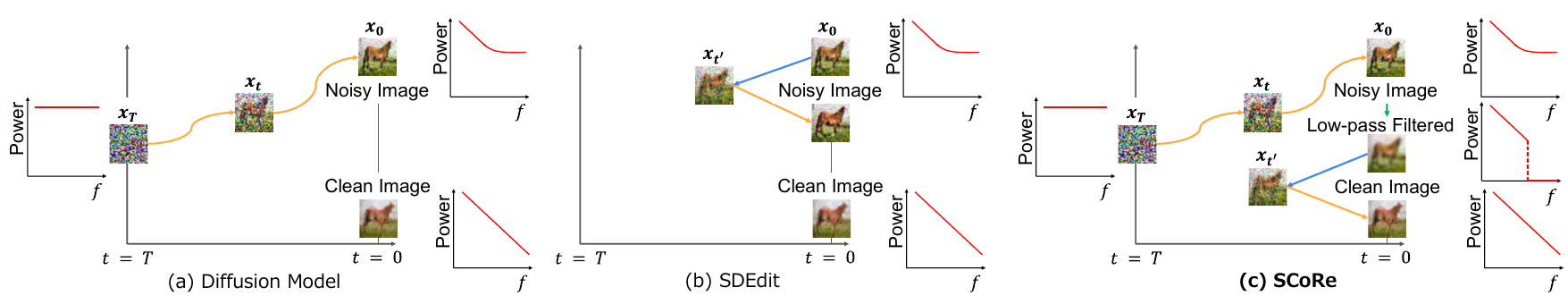}\\[-4mm]
    \caption{The generation process of a diffusion model trained on noisy images. (a)~Standard generation process: produces noisy generated images. (b)~SDEdit: When SDEdit is applied to a noisy generated image, the result is regenerated as a noisy image. (c)~SCoRe: Removal of high-frequency components via a cutoff operation, followed by regeneration using SDEdit to yield cleaner images.}
    \label{method_}
    \vspace{-5pt}
\end{figure*}

\section{SCoRe: Spectral Cutoff Regeneration}
Our proposed method, SCoRe (Spectral Cutoff Regeneration), addresses the problem that diffusion models trained on datasets containing noisy images can reflect the influence of noise in their samples, particularly in high-frequency components. Without any additional training, SCoRe performs generation-time spectral regeneration: it first creates a low-pass filtered image by suppressing frequency components above a cutoff frequency, then injects an appropriate amount of noise through the diffusion process, and finally regenerates high-frequency details via the reverse process (SDEdit). The amount of injected noise (i.e., the SDEdit initialization timestep) is determined from the chosen cutoff frequency via RAPSD-based analysis, which avoids excessive noise injection and prevents the noise influence from being reintroduced during regeneration.

\subsection{Diffusion Models}
Diffusion models~\cite{diffusionmodel,DDPM} consist of a \emph{diffusion process} that adds noise to data and a \emph{generation process} (reverse process) that gradually removes the noise and brings samples closer to the data distribution. In the diffusion process, Gaussian noise is progressively added to an image \(x_0\) sampled from the dataset according to the timestep, until the image is transformed into pure random noise. The marginal distribution of \(x_t\) at timestep \(t\) can be written in closed form as
\begin{equation}
    x_t = \sqrt{\bar{\alpha}_t}\,x_0 + \sqrt{1-\bar{\alpha}_t}\,\epsilon,
    \label{diff_process}
\end{equation}
where \(\epsilon \sim \mathcal{N}(0,I)\) and \(\bar{\alpha}_t = \prod_{s=1}^{t}(1-\beta_s)\) is determined by a noise schedule \(\{\beta_t\}\).

In the reverse process, an image is generated by gradually removing noise starting from random noise. As illustrated in Fig.~\ref{method_}~(a), the reverse process starts from a noisy sample \(x_T\) and iteratively obtains \(x_0\) by removing the predicted noise component at each step. For an image \(x_t\) at timestep \(t\), using the noise estimate \(\epsilon_\theta(x_t,t)\) predicted by a model \(\epsilon_\theta\) with parameters \(\theta\), the denoised sample \(x_{t-1}\) is given by
\begin{equation}
    x_{t-1} = \frac{1}{\sqrt{1 - \beta_t}} 
    \left( x_t - \frac{\beta_t}{\sqrt{1 - \bar{\alpha}_t}} \, \epsilon_\theta(x_t, t) \right) 
    + \sigma_t z,
    \label{revrse_process}
\end{equation}
where \(\sigma_t = \sqrt{\beta_t}\) and \(z \sim \mathcal{N}(0, I)\).
The model \(\epsilon_\theta\) is trained to predict the noise \(\epsilon\) added to \(x_t\) at timestep \(t\) by minimizing the following loss function \(\mathcal{L}_{\mathrm{simple}}\):
\begin{equation}
    \mathcal{L}_{\mathrm{simple}} = \mathbb{E}_{x_0, t, \epsilon} \left[ \|\epsilon - \epsilon_\theta(x_t, t) \|^2 \right].
\end{equation}

\subsection{SDEdit}
SDEdit~\cite{SDEdit} is an image editing method based on diffusion models. First, given an input image $x$, Gaussian noise is added according to the diffusion process up to timestep $t'(<T)$ to obtain a diffused image $x_{t'}$. This diffusion process is expressed as:
\begin{equation}
x_{t'} = \sqrt{\bar{\alpha}_{t'}}\,x + \sqrt{1-\bar{\alpha}_{t'}}\,\epsilon,
\quad \epsilon \sim \mathcal{N}(0,I)
\label{SDEdit_diff_process}
\end{equation}
Then, starting from the diffused image $x_{t'}$, a new image is generated by applying the reverse process from $t'$ down to $0$ (i.e., for $t'$ steps).

\subsection{Proposed Method}
Our method, SCoRe, performs training-free, generation-time spectral regeneration to produce cleaner samples from diffusion models trained on datasets containing a mixture of clean and noisy images. Leveraging the tendency of diffusion models to infer high-frequency details from low-frequency structure, SCoRe suppresses corrupted high-frequency components via a cutoff operation and then regenerates high-frequency details via SDEdit, without any retraining or fine-tuning.

In the standard sampling procedure of diffusion models, a generated image \(x_0\) is obtained by running the reverse process starting from pure noise \(x_T\). However, due to degradations present in the training data, \(x_0\) can become a noisy generated image that contains noise in its high-frequency components as shown in Fig.~\ref{method_}~(a). Moreover, even if SDEdit is naively applied to such a noisy generated image, the influence of noise tends to remain, and the generation results are not sufficiently improved as shown in Fig.~\ref{method_}~(b).

We focus on the statistical properties of images in the frequency domain to suppress residual noise in generated results while encouraging the regeneration of high-frequency components.
Typically, low-frequency components dominate images, whereas high-frequency components mainly represent fine details. However, degradation artifacts in generated results may also appear in the high-frequency bands.
To address this issue, we apply SDEdit to an input image $x_\mathrm{cutoff}$ obtained by suppressing high-frequency $f_\mathrm{cutoff}$ components via a cutoff operation, and then regenerate the image through the re-diffusion and reverse processes (Fig.~\ref{method_}~(c)).
During SDEdit, high-frequency components are progressively regenerated as the reverse proceeds. Setting the diffusion timestep $t'$ to the value determined by Eq.~8 prevents the noise power in the high-frequency components from increasing excessively, thereby encouraging the addition of fine details consistent with the model prior.
The re-diffusion process in SDEdit is expressed as follows:
\begin{equation}
     x_{t'} = \sqrt{\bar{\alpha}_{t'}}\,x_\mathrm{cutoff}
     + \sqrt{1-\bar{\alpha}_{t'}}\,\epsilon,
     \quad \epsilon \sim \mathcal{N}(0,I)
     \label{proposed_diff_process}
\end{equation}

\subsection{Relationship between The Cutoff Frequency and The SDEdit Timestep $t'$}
We use, as input to SDEdit, an image whose high-frequency components above frequency $f_\mathrm{cutoff}$ are removed by a cutoff operation. This means that the diffusion model is tasked with regenerating the frequency components higher than the cutoff frequency. As described above, diffusion models estimate image content progressively from low- to high-frequency components. Therefore, ideally, SDEdit should be initialized at a timestep corresponding to the cutoff frequency.

Accordingly, for a given cutoff frequency $f_\mathrm{cutoff}$, we determine the corresponding diffusion timestep $t'$ for SDEdit. We formulate this correspondence by analyzing the diffusion process via Radially Averaged Power Spectral Density (RAPSD). Specifically, given a set of $N$ training images $\{x_0^i\}_{i=1}^{N}$, we define $P_t(f)$ as the average power at frequency $f$ computed from the RAPSD of the diffused images $\{x_t^i\}_{i=1}^{N}$ at timestep $t$. Moreover, by using this function and Eq.~(\ref{diff_process}), we can define the Signal-to-Noise Ratio (SNR) in the diffusion process at timestep $t$ as:
\begin{equation}\mathrm{SNR}_{t}(f)=\frac{\bar{\alpha}_{t}P_0(f)}{(1-\bar{\alpha}_{t})P_T(f)},\end{equation}
where $P_0(f)$ and $P_T(f)$ represent the power spectra of the clean data and the random noise, respectively.

Based on the SNR, we define the diffusion timestep $t'$ for the cutoff frequency $f_\mathrm{cutoff}$ as the step satisfying $\mathrm{SNR}_{t'}(f_\mathrm{cutoff})=1$. Because this condition represents the crossover point where the signal power equals the noise power, we consider it the ideal boundary for switching between signal preservation and regeneration. Consequently, we can derive the following equation:
\begin{equation}\frac{\bar{\alpha}_{t'}}{1-\bar{\alpha}_{t'}} = \frac{P_T(f_\mathrm{cutoff})}{P_0(f_\mathrm{cutoff})}.\label{eq:boundary}
\end{equation}
By using this equation and the noise schedule $\{\bar{\alpha}_{t}\}_{t=1}^{T}$, we can finally determine the diffusion timestep $t'$ as follows:
\begin{equation}t' = \bar{\alpha}_t^{-1}\left(\frac{P_T(f_\mathrm{cutoff})}{P_0(f_\mathrm{cutoff})+P_T(f_\mathrm{cutoff})}\right).\label{eq:t_star}
\end{equation}

\section{Experimental results}
We evaluated the effectiveness of the proposed method under two settings: synthetic noisy images and real-world noisy images. First, using the CIFAR-10 dataset with artificially added noise (synthetic noise), we analyzed the basic characteristics of the proposed method and its behavior under different noise types. Next, considering practical applications, we conducted experiments on the SIDD dataset, which contains real-world imaging noise, to evaluate the robustness and practicality of the proposed method against complex and unknown noise.

\begin{figure*}[t]
    \centering
    \includegraphics[width=\linewidth]{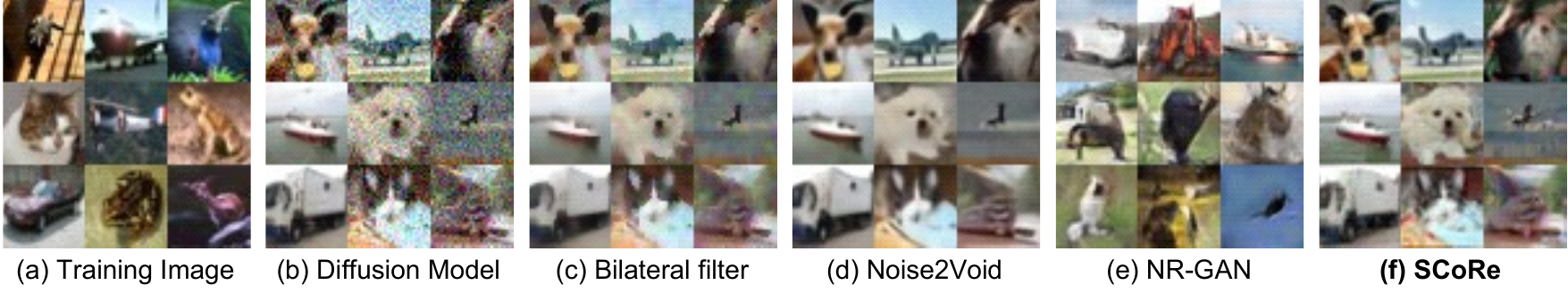}\\[-4mm]
    \caption{Generated images under CIFAR-10 with Gaussian noise: (a) Training examples, (b) standard diffusion sampling, (c) generated images post-processed with a bilateral filter, (d) generated images post-processed with Noise2Void, (e) NR-GAN, and (f) SCoRe.}
    \label{cifar_generated_images}
\end{figure*}

\subsection{Synthetic Noisy Images}
\noindent\textbf{Dataset containing noisy training images:}
To train the diffusion model, we used the 50,000 training images of CIFAR-10 to construct a training dataset where 10\% of the images were clean and the remaining 90\% were treated as noisy training images. The noisy training images were created by adding noise to the clean training images. Following prior work~\cite{NR-GAN}, we considered three types of noise:
(1)~Gaussian: Gaussian noise with mean 0 and standard deviation 25;
(2)~Poisson: noise sampled from a Poisson distribution with mean \(30\) and variance \(30\) (i.e., \(\lambda=30\));
(3)~Mix: noise obtained by first adding Poisson noise and then adding Gaussian noise.

\noindent\textbf{Comparison methods:}
As comparison methods, we used a standard diffusion model (Diffusion Model) as well as approaches that apply denoising to the generated images after sampling. For post-processing, we adopted a bilateral filter and evaluated a method that applies the filter as post-processing to the generated images (Post-Bilateral). In addition, we included Noise2Void (N2V), a denoising model, by applying it as post-processing in the same manner (Post-N2V). We also applied standard SDEdit to images generated by the standard diffusion model (SDEdit) to assess the effect of the cutoff operation in the proposed method. Moreover, we applied NR-GAN~\cite{NR-GAN}, a learning-based method designed to generate images from datasets containing noisy training images. For all methods, including the proposed method, we trained the diffusion model for approximately 800K iterations with a batch size of 128, and trained NR-GAN according to its original settings.

\noindent\textbf{Metrics:}
To evaluate the quality of generated images, we used the Fr\'echet Inception Distance (FID)~\cite{FID}. The FID measures the distance between the distribution of generated images and that of real images, where lower values indicate higher generation quality.
For evaluation, we computed FID between 50,000 clean real images and 50,000 generated images using the InceptionV3 network pre-trained on ImageNet~\cite{InceptionV3}.

\begin{figure}[t]
    \centering
    \includegraphics[width=0.70\columnwidth]{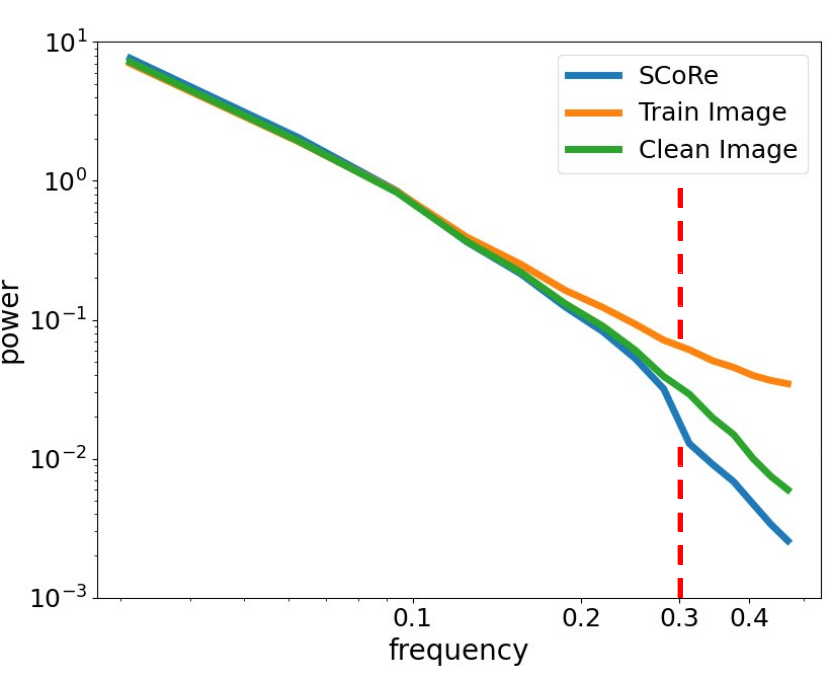}\\[-4mm]
    \caption{Comparison of the RAPSDs of clean images, training images, and the results produced by the proposed method under Gaussian noise.}
    \label{cifar_RAPSD}
\end{figure}

\noindent\textbf{Cutoff frequency settings:}
In this experimental setting, we set the default cutoff frequency to $f_\mathrm{cutoff}=0.30$ based on the power of the injected noise across frequency bands. In practical deployment, however, the noise characteristics may be unknown and the optimal cutoff value cannot be determined in advance. Therefore, to examine whether the proposed method works robustly without being overly sensitive to the cutoff choice, we also evaluate nearby values $f_\mathrm{cutoff}=0.20$ and $f_\mathrm{cutoff}=0.40$. The corresponding SDEdit diffusion timestep $t'$ for each cutoff frequency is determined according to Eq.~(\ref{eq:t_star}).

Furthermore, in Table~\ref{tab:cifar_FID_results} and Table~\ref{tab:SIDD_results}, we denote our method as \(\mathrm{SCoRe}(f_\mathrm{cutoff},t')\), where $f_\mathrm{cutoff}$ is the cutoff frequency and \(t'\) is the SDEdit diffusion timestep determined from $f_\mathrm{cutoff}$ according to Eq.~(\ref{eq:t_star}). For example, \(\mathrm{SCoRe}(0.30, 72)\) indicates that we use a cutoff frequency of $f_\mathrm{cutoff}=0.30$ and set the corresponding timestep to \(t'=72\).

\begin{table}[t]
    \centering
    \caption{Comparison of FID on synthetic noisy dataset}
    \vspace{5pt}
    \renewcommand{\arraystretch}{1.3}
    \begin{tabular}{l|ccc}
        \toprule
        Method & Gaussian & Poisson & Mix \\ 
        \hline
        Diffusion Model  & 132.9 & 168.7 & 203.2 \\ 
        \hspace{10pt}w/ Post-Bilateral   & 51.7 & 68.0 & 102.8 \\
        \hspace{10pt}w/ Post-N2V & 32.3 & \underline{31.6} & 40.1 \\ 
        SDEdit & 132.2 & 151.3 & 180.2 \\ 
        NR-GAN   & \underline{15.5} & 46.5 & \underline{32.0} \\
        \rowcolor{gray!15}\textbf{SCoRe}\ ($f_\mathrm{cutoff}=0.20$, $t'=119$) & 14.0 & 14.9 & \textbf{16.2} \\
        \rowcolor{gray!15}\textbf{SCoRe}\ ($f_\mathrm{cutoff}=0.30$, $t'=72$)  & \textbf{9.8} & \textbf{14.7} & 25.8 \\
        \rowcolor{gray!15}\textbf{SCoRe}\ ($f_\mathrm{cutoff}=0.40$, $t'=58$)  & 11.0 & 24.2 & 45.7 \\

        \bottomrule
    \end{tabular}
    \label{tab:cifar_FID_results}
\end{table}

\noindent\textbf{Quantitative results:}\ 
Table~\ref{tab:cifar_FID_results} shows the FID scores. The standard diffusion model suffers from degraded quality due to the noisy training data. Post-processing with a bilateral filter improves performance, particularly for Gaussian noise, while Noise2Void offers consistent improvements across all conditions. In contrast, naive SDEdit yields negligible changes as it tends to regenerate the input noise. NR-GAN outperforms the standard model but struggles with Poisson and Mix noise, likely due to inaccurate noise estimation. Significantly, the proposed method achieves the best FID scores under all conditions. This confirms that our frequency-based strategy effectively removes noise while regenerating clean details. Furthermore, our method outperforms baselines even with sub-optimal cutoff values, demonstrating its robustness.

\noindent\textbf{Qualitative results:}\ Fig.~\ref{cifar_generated_images} presents the generated samples under the Gaussian noise condition. As observed in Fig.~\ref{cifar_generated_images}~(b), the standard diffusion model fails to suppress the noise inherited from the training data. While the (c) bilateral filter removes noise, it inadvertently blurs clean image components. Similarly, (d) Noise2Void and (e) NR-GAN yield unsatisfactory results, suffering from monotonous textures and low object fidelity, respectively. In contrast, (f) the proposed method successfully generates clean images with high fidelity, effectively removing noise while preserving structural details.

\noindent\textbf{Frequency analysis:}
Fig.~\ref{cifar_RAPSD} shows the mean RAPSD. Compared with training images, the proposed method yielded spectral magnitudes in the mid-to-high frequency bands that were closer to those of the ground-truth images, indicating improved high-frequency components. On the other hand, in the high-frequency band, the proposed method also showed a slightly lower spectral magnitude than the clean (ground-truth) images. However, since natural images generally have much larger power in low-frequency components, differences in the high-frequency band are relatively small, and thus their impact on the perceptual quality of the generated images is limited.

\begin{figure}[t]
    \centering
    \includegraphics[width=0.70\linewidth]{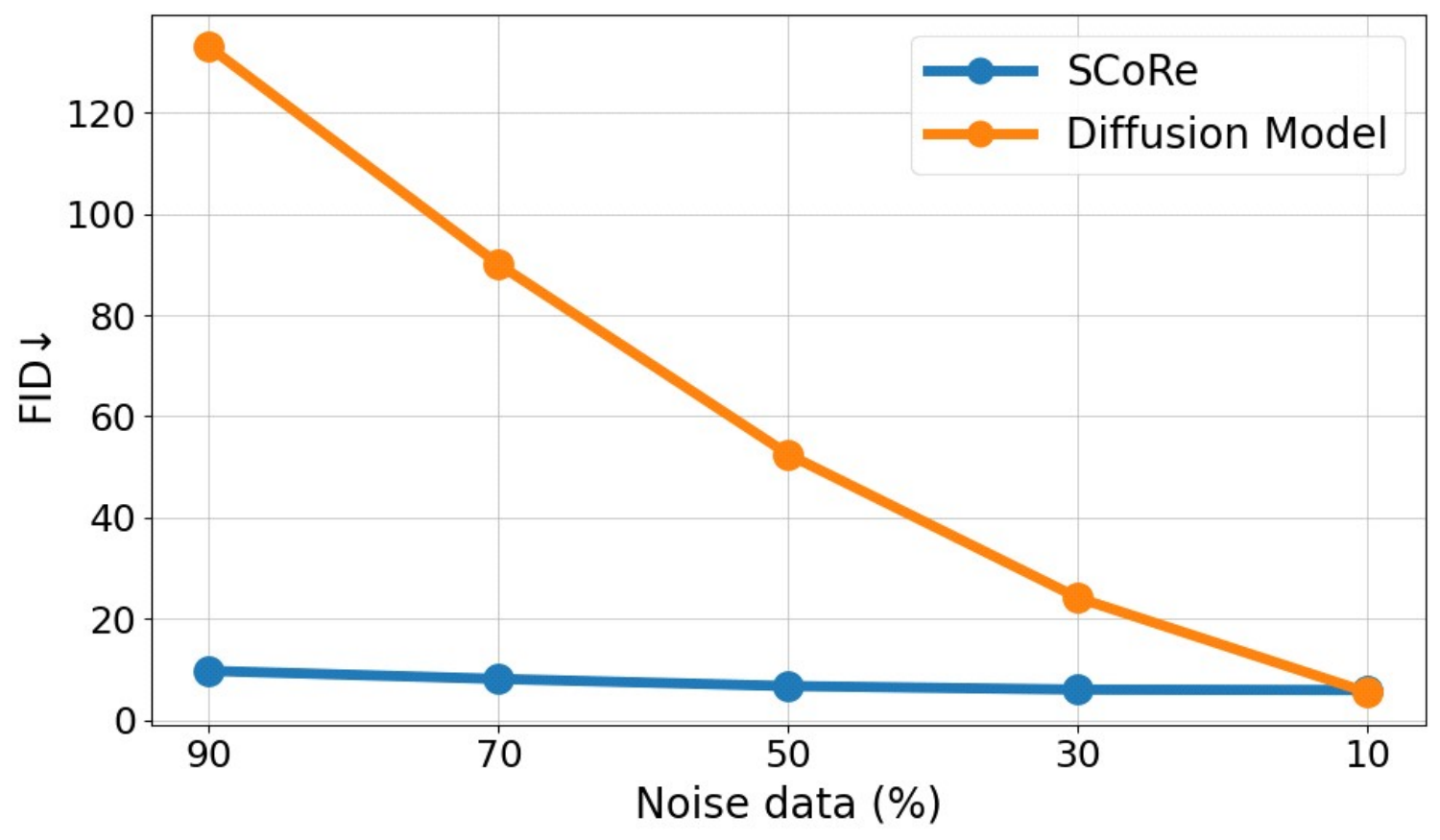}\\[-4mm]
    \caption{Effect of the noisy-data ratio in training on FID (x-axis: percentage of noisy training images).}
    \label{ratio_cifar}
\end{figure}

\noindent\textbf{Analysis of the ratio of noisy and clean samples:}\ 
We further investigated the impact of the noisy data proportion on generative performance by varying the ratio of noisy images from 10\% to 90\%. Fig.~\ref{ratio_cifar} illustrates the results. The standard diffusion model shows a linear deterioration in FID as the noise ratio increases. In sharp contrast, the proposed method consistently maintains low FID scores across all noise levels, demonstrating exceptional robustness to data contamination.In low-noise regimes (10\%), the gap narrows, but our method maintains stability and quality comparable to the baseline, demonstrating effectiveness regardless of the noise ratio.

\subsection{Real-world Noisy Images}
\noindent\textbf{Dataset of real-world noise: }\ 
We used the SIDD~\cite{SIDD} dataset. The dataset consists of real photographs captured with smartphones and contains realistic imaging noise in which signal-dependent and signal-independent components are mixed, such as shot noise and signal-independent additive noise (e.g., read-noise-like components). For dataset construction, we cropped the images in the original dataset into $128\times128$ patches and built a training set of 8,160 cropped images. The training set was then composed such that 90\% of the images were noisy and 10\% were clean.

\begin{figure*}[t]
    \centering
    \includegraphics[width=\linewidth]{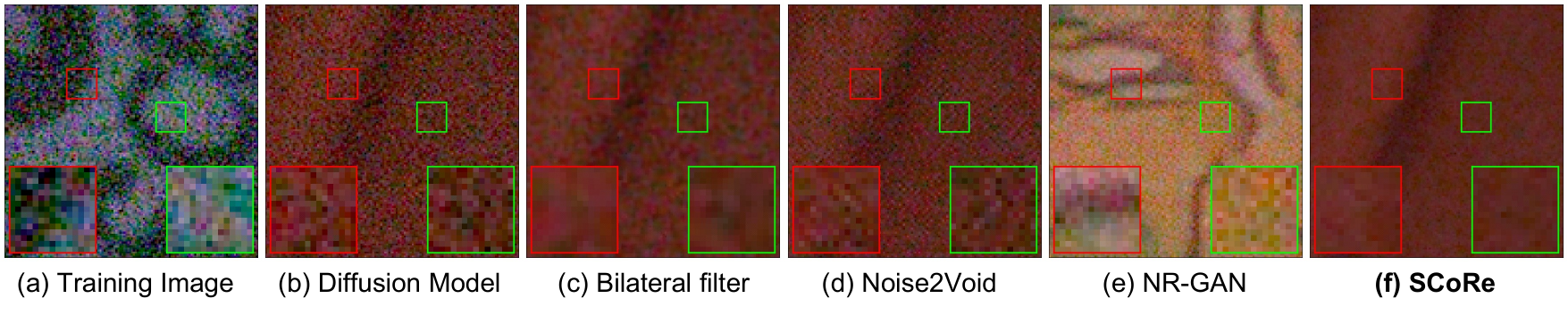}
    \caption{Generated Results : (a) training examples, (b) standard diffusion sampling, (c) generated images post-processed with a bilateral filter, (d) generated images post-processed with Noise2Void, (e) NR-GAN, and (f) SCoRe.}
    \label{SIDD_generated_images}
\end{figure*}

\noindent\textbf{Other settings:}\ For all methods, we trained the diffusion model for approximately 520K iterations with a batch size of 64. Moreover, in our noise setting, we set $f_\mathrm{cutoff}=0.10$ as a reference cutoff value. 
For FID evaluation, we use 8,160 clean images and the corresponding generated images.

\noindent\textbf{Quantitative results:}\ 
Table~\ref{tab:SIDD_results} presents the FID scores. The standard diffusion model suffers from quality degradation due to the noisy training data. Applying a bilateral filter worsens the FID, likely due to a mismatch with the complex noise characteristics of SIDD. While Noise2Void~\cite{Noise2Void} improves performance, naive SDEdit shows negligible change as it tends to regenerate the input noise. Similarly, NR-GAN~\cite{NR-GAN} performs worse than the baseline, presumably because its generator fails to accurately model real-world noise distributions. In contrast, the proposed method significantly outperforms all other baselines. This demonstrates that our method can effectively generate clean images even when trained on data corrupted by real-world noise.

\noindent\textbf{Qualitative results:}\ 
Fig.~\ref{SIDD_generated_images} presents the generated results. As shown in Fig.~\ref{SIDD_generated_images}~(b), the standard diffusion model still produces images with noticeable noise. In Fig.~\ref{SIDD_generated_images}~(c), residual noise is observed and fine details are blurred. Similarly, Fig.~\ref{SIDD_generated_images}~(d) also shows remaining noise. In Fig.~\ref{SIDD_generated_images}~(e), noise persists, indicating that the model fails to generate clean images. In contrast, Fig.~\ref{SIDD_generated_images}~(f) demonstrates that noise is effectively reduced and faithful images are regenerated without blurring. These results confirm that the proposed method can generate clean images even when using a diffusion model trained on a dataset containing real-world noisy images.

\begin{table}[t]
    \centering
    \caption{Comparison of FID on real-world noisy dataset}
    \vspace{5pt}
    \renewcommand{\arraystretch}{1.3}
    \begin{tabular}{l|ccc}
        \toprule
        Method & SIDD \\ 
        \hline
        Diffusion Model  & 30.1 \\ 
        \hspace{10pt}w/ Post-Bilateral & 30.9 \\
        \hspace{10pt}w/ Post-N2V & \underline{27.0} \\
        SDEdit & 30.6 \\
        NR-GAN   & 46.3  \\
        \rowcolor{gray!15}\textbf{SCoRe}\ ($f_\mathrm{cutoff}=0.05$, $t'=110$) & 29.0  \\
        \rowcolor{gray!15}\textbf{SCoRe}\ ($f_\mathrm{cutoff}=0.10$, $t'=63$)  & 16.8  \\
        \rowcolor{gray!15}\textbf{SCoRe}\ ($f_\mathrm{cutoff}=0.15$, $t'=45$)  & \textbf{16.2}  \\
    \bottomrule
    \end{tabular}
    \label{tab:SIDD_results}
\end{table}

\section{Conclusion}
In this work, we address clean image generation from diffusion models trained on noisy datasets. We propose SCoRe (Spectral Cutoff Regeneration), a training-free generation-time method that leverages the spectral bias of diffusion models to regenerate high-frequency details from preserved low-frequency structures. By deriving the SDEdit initialization timestep via RAPSD analysis, SCoRe suppresses noise artifacts without additional training. Experiments on synthetic and real-world datasets show consistent improvements over existing methods.

\section*{ACKNOWLEDGMENT}
This work was supported by JSPS KAKENHI Grant Number JP22H05180 and JP24KJ180.

\bibliographystyle{IEEEtran}
\bibliography{refs}

\end{document}